%% file: 2016-PGQL-shiny.tex
\documentclass{article}

\usepackage[dvipsnames]{xcolor}
\usepackage[left=1in,right=1in,top=1in,bottom=1in]{geometry}
\usepackage{hyperref,natbib}
\hypersetup{
colorlinks   = true, 
urlcolor     = blue, 
linkcolor    = blue, 
citecolor   = black 
}
\usepackage{url}
\usepackage{amsmath}
\usepackage{amssymb,mathtools,amsthm}
\usepackage{bm}
\usepackage{mdframed}
\usepackage{subfigure}

\input{rl_math}
\usepackage{cleveref}

\newcommand{\titleQ}{\texorpdfstring{$Q$}{Q}}

\crefname{prop}{proposition}{propositions}

\newcommand{\Qtarg}{\underline{Q}}

\newcommand{\pith}{\pi_{\theta}}
\newcommand{\gradth}{\grad_{\theta}}

\newcommand{\Qpigam}{Q^{\pi,\gamma}}

\newcommand{\hatv}{\hat{V}}
\newcommand{\hatq}{\hat{Q}}

\newcommand{\temp}{\tau}
\newcommand{\KL}{\operatorname{KL}}
\newcommand{\T}{\mathcal{T}}
\newcommand{\Tpi}{\mathcal{T}_{\pi}}

\newcommand{\where}{\quad\text{where}\quad}

\newcommand{\Qth}{Q_{\theta}}

\newcommand{\pistar}{\pi^{\ast}}

\newcommand{\Vth}{V_{\theta}}
\newcommand{\Dlam}{\Delta}
\newcommand{\bPsi}{\bm{\Psi}}
\newcommand{\bpsi}{\bm{\psi}}

\newcommand{\bw}{\mathbf{w}}
\newcommand{\pithold}{\pi_{\theta_{\text{old}}}}
\newcommand{\Qthold}{Q_{\theta_{\text{old}}}}
\newcommand{\Vthold}{V_{\theta_{\text{old}}}}
\newcommand{\hatqeps}{\hatq^{\epsilon}}

\newcommand{\thold}{\theta_{\text{old}}}
\newcommand{\G}{\mathcal{G}}
\newcommand{\ggam}{g_{\gamma}}
\newcommand{\K}{\KL}
\newcommand{\piref}{\overline{\pi}}
\newcommand{\expf}[1]{\exp(#1)}

\newcommand{\ptfrac}[2]{\lrparen*{\tfrac{#1}{#2}}}

\newcommand{\relent}{\kl{\pith}{\piref}}
\newcommand{\pifrac}{\tfrac{\pi(a_t \given s_t)}{\piref(a_t \given s_t)}}
\newcommand{\barr}{\bar{r}}
\newcommand{\qth}{q_{\theta}}

\newcommand{\vth}{v_{\theta}}
\newcommand{\pibol}{\pi^{\mathcal{B}}}
\newcommand{\pibolq}{\pibol_Q}
\newcommand{\pibolth}{\pibol_{\qth}}

\newcommand{\belowcomment}[1]{ &\qquad\qquad\text{\color{RoyalPurple} #1} \nonumber \\ }

\title{Equivalence Between Policy Gradients and Soft Q-Learning}

\begin{document}

\begin{center}
{\Large \textbf{Equivalence Between Policy Gradients and Soft $Q$-Learning}}\\
\vspace{1em}
{\large John Schulman$^{1}$, Xi Chen$^{1,2}$, and Pieter Abbeel$^{1,2}$}\\
\vspace{0.5em}
$^{1}$OpenAI \qquad $^{2}$UC Berkeley, EECS Dept.\\
\vspace{0.5em}
\textit{\{joschu, peter, pieter\}@openai.com}
\end{center}

\begin{abstract}
Two of the leading approaches for model-free reinforcement learning are policy gradient methods and $Q$-learning methods.
$Q$-learning methods can be effective and sample-efficient when they work, however, it is not well-understood why they work, since empirically, the $Q$-values they estimate are very inaccurate.
A partial explanation may be that $Q$-learning methods are secretly implementing policy gradient updates: we show that there is a precise equivalence between $Q$-learning and policy gradient methods in the setting of entropy-regularized reinforcement learning, that ``soft'' (entropy-regularized) $Q$-learning is exactly equivalent to a policy gradient method.
We also point out a connection between $Q$-learning methods and natural policy gradient methods.

Experimentally, we explore the entropy-regularized versions of $Q$-learning and policy gradients, and we find them to perform as well as (or slightly better than) the standard variants on the Atari benchmark.
We also show that the equivalence holds in practical settings by constructing a $Q$-learning method that closely matches the learning dynamics of A3C without using a target network or $\epsilon$-greedy exploration schedule.

\end{abstract}

\section{Introduction}
Policy gradient methods (PG) and $Q$-learning (QL) methods perform updates that are qualitatively similar.
In both cases, if the return following an action $a_t$ is high, then that action is reinforced:
in policy gradient methods, the probability $\pi(a_t \given s_t)$ is increased; whereas in $Q$-learning methods, the $Q$-value $Q(s_t,a_t)$ is increased.
The connection becomes closer when we add entropy regularization to these algorithms.
With an entropy cost added to the returns, the optimal policy has the form $\pi(a \given s) \propto \exp(Q(s,a))$; hence policy gradient methods solve for the optimal $Q$-function, up to an additive constant (\cite{ziebart2010modeling}).
\cite{o2016pgq} also discuss the connection between the fixed points and updates of PG and QL methods, though the discussion of fixed points is restricted to the tabular setting, and the discussion comparing updates is informal and shows an approximate equivalence.
Going beyond past work, this paper shows that under appropriate conditions, the gradient of the loss function used in $n$-step $Q$-learning is equal to the gradient of the loss used in an $n$-step policy gradient method, including a squared-error term on the value function.
Altogether, the update matches what is typically done in ``actor-critic'' policy gradient methods such as A3C, which explains why \citet{mnih2016asynchronous} obtained qualitatively similar results from policy gradients and $n$-step $Q$-learning.

\Cref{secbandit} uses the bandit setting to provide the reader with a simplified version of our main calculation. (The main calculation applies to the MDP setting.)
\Cref{secerrl} discusses the entropy-regularized formulation of RL, which is not original to this work, but is included for the reader's convenience.
\Cref{segpg} shows that the soft $Q$-learning loss gradient can be interpreted as a policy gradient term plus a baseline-error-gradient term, corresponding to policy gradient instantiations such as A3C \citep{mnih2016asynchronous}.
\Cref{secnpg} draws a connection between QL methods that use batch updates or replay-buffers, and natural policy gradient methods.

Some previous work on entropy regularized reinforcement learning (e.g., \cite{o2016pgq,nachum2017bridging}) uses entropy bonuses, whereas we use a penalty on Kullback-Leibler (KL) divergence, which is a bit more general. However, in the text, we often refer to ``entropy'' terms; this refers to ``relative entropy'', i.e., the KL divergence.

\section{Bandit Setting} \label{secbandit}

Let's consider a bandit problem with a discrete or continuous action space: at each timestep the agent chooses an action $a$, and the reward $r$ is sampled according to $P(r \given a)$, where $P$ is unknown to the agent.
Let $\barr(a) = \Ea{r \given a}$, and let $\pi$ denote a policy, where $\pi(a)$ is the probability of action $a$.
Then, the expected per-timestep reward of the policy $\pi$ is $\Eb{a \sim \pi}{r} = \sum_a \pi(a) \barr(a)$ or $\myint{a} \pi(a) \barr(a)$.
Let's suppose we are maximizing $\eta(\pi)$, an entropy-regularized version of this objective:
\begin{align}
\eta(\pi) = \Eb{a\sim\pi, r}{r} - \tau \kl{\pi}{\piref}
\end{align}
where $\piref$ is some ``reference'' policy, $\tau$ is a ``temperature'' parameter, and $D_{\KL}$ is the Kullback-Leibler divergence.
Note that the temperature $\tau$ can be eliminated by rescaling the rewards. However, we will leave it so that our calculations are checkable through dimensional analysis, and to make the temperature-dependence more explicit.

First, let us calculate the policy $\pi$ that maximizes $\eta$.
We claim that $\eta(\pi)$ is maximized by $\pibol_{\barr}$, defined as
\begin{align}
\pibol_{\barr}(a) = \piref(a) \expf{\barr(a)/\tau} / \underbrace{\Eb{a' \sim \piref}{\expf{\barr(a')/\tau}}}_{\text{normalizing constant}}.    \label{pistardef}
\end{align}
To derive this, consider the KL divergence between $\pi$ and $\pibol_{\barr}$:
\begin{align}
\kl{\pi}{\pibol_{\barr}}
&= \Eb{a \sim \pi}{\log \pi(a) - \log \pibol_{\barr}(a)}\\
&= \Eb{a \sim \pi}{\log \pi(a) - \log \piref(a) - \barr(a)/\tau + \log \Eb{a \sim \piref}{\expf{\barr(a)/\tau}}}\\
&= \kl{\pi}{\piref} - \Eb{a \sim \pi}{\barr(a)/\tau} + \log \Eb{a \sim \piref}{\expf{\barr(a)/\tau}}
\end{align}
Rearranging and multiplying by $\tau$,
\begin{align}
\Eb{a \sim \pi}{\barr(a)} - \tau \kl{\pi}{\piref} = \tau \log \Eb{a \sim \piref}{\expf{\barr(a)/\tau}} - \tau \kl{\pi}{\pibol_{\barr}} \label{kl-ident}
\end{align}
Clearly the left-hand side is maximized (with respect to $\pi$) when the KL term on the right-hand side is minimized (as the other term does not depend on $\pi$), and $\kl{\pi}{\pibol_{\barr}}$ is minimized at $\pi=\pibol_{\barr}$.

The preceding calculation gives us the optimal policy when $\barr$ is known, but in the entropy-regularized bandit problem, it is initially unknown, and the agent learns about it by sampling.
There are two approaches for solving the entropy-regularized bandit problem:
\begin{enumerate}
\item A direct, policy-based approach, where we incrementally update the agent's policy $\pi$ based on stochastic gradient ascent on $\eta$.
\item An indirect, value-based approach, where we learn an action-value function $\qth$ that estimates and approximates $\barr$, and we define $\pi$ based on our current estimate of $\qth$.
\end{enumerate}
For the policy-based approach, we can obtain unbiased estimates the gradient of $\eta$. For a parameterized policy $\pith$, the gradient is given by
\begin{align}
\gradth \eta(\pith) = \Eb{a \sim \pith, r}{\gradth \log \pith(a) r - \tau \gradth \kl{\pith}{\piref}}. \label{bandit-pg}
\end{align}
We can obtain an unbiased gradient estimate using a single sample $(a,r)$.

In the indirect, value-based approach approach, it is natural to use a squared-error loss:
\begin{align}
L_{\pi}(\theta)  \defeq \half\Eb{a \sim \pi, r}{\lrparen*{\qth(a) - r}^2}
\end{align}
Taking the gradient of this loss, with respect to the parameters of $\qth$, we get
\begin{align}
\gradth L_{\pi}(\theta) = \Eb{a \sim \pi, r}{\gradth \qth(a) (\qth(a) - r)} \label{bandit-vlossgrad}
\end{align}
Soon, we will calculate the relationship between this loss gradient and the policy gradient from \Cref{bandit-pg}.

In the indirect, value-based approach, a natural choice for policy $\pi$ is the one that would be optimal if $\qth = \barr$.
Let's denote this policy, called the Boltzmann policy, by $\pibolth$, where
\begin{align}
\pibolth(a) = \piref(a) \exp(\qth(a) / \tau) / \Eb{a' \sim \piref}{\exp(\qth(a')/\tau)}.
\end{align}
It will be convenient to introduce a bit of notation for the normalizing factor; namely, we define the scalar
\begin{align}
\vth = \tau \log \Eb{a \sim \piref}{\exp(\qth(a))/\tau} \label{banditvdef}
\end{align}
Then the Boltzmann policy can be written as
\begin{align}
\pibolth(a) = \piref(a) \exp((\qth(a) - \vth)/\tau). \label{bandit-softmax}
\end{align}
Note that the term $\tau \log \Eb{a \sim \piref}{\expf{\barr(a)/\tau}}$, appeared earlier in \Cref{kl-ident}). Repeating the calculation from \Cref{pistardef} through \Cref{kl-ident}, but with $\qth$ instead of $\barr$,
\begin{align}
\vth =  \Eb{a \sim \pibolth}{\qth(a)} - \tau \kl{\pibolth}{\piref}. \label{banditvthid}
\end{align}
Hence, $\vth$ is an estimate of $\eta(\pibolth)$, plugging in $\qth$ for $\barr$.

Now we shall show the connection between the gradient of the squared-error loss (\Cref{bandit-vlossgrad}) and the policy gradient (\Cref{bandit-pg}).
Rearranging \Cref{bandit-softmax}, we can write $\qth$ in terms of $\vth$ and the Boltzmann policy $\pibolth$:
\begin{align}
\qth(a) &= \vth + \tau \log \ptfrac{\pibolth(a)} {\piref(a)}
\end{align}
Let's substitute this expression for $\qth$ into the squared-error loss gradient (\Cref{bandit-vlossgrad}).
\begin{align}
&\gradth L_{\pi}(\qth)
= \Eb{a \sim \pi,r}{\gradth \qth(a) (\qth(a) - r)} \\
&= \Eb{a \sim \pi,r}{\gradth\lrparen*{\vth + \tau \log \ptfrac{\pibolth(a)}{\piref(a)}} \lrparen*{\vth + \tau \log \ptfrac{\pibolth(a)}{\piref(a)} - r}}\\
&= \Eb{a \sim \pi,r}{\tau \gradth \log\pibolth(a) \lrparen*{\vth + \tau \log \ptfrac{\pibolth(a)}{\piref(a)} - r} + \gradth \vth \lrparen*{\vth + \tau \log \ptfrac{\pibolth(a)}{\piref(a)}-r}} \label{banditpart1end}
\end{align}
Note that we have not yet decided on a sampling distribution $\pi$. Henceforth, we'll assume actions were sampled by $\pi=\pibolth$.
Also, note the derivative of the KL-divergence:
\begin{align}
\gradth \kl{\pibolth}{\piref}
&= \gradth \myint{a} \pibolth(a) \log \ptfrac{\pibolth(a)}{\piref(a)}\\
&= \myint{a} \gradth \pibolth(a)\lrparen*{ \log \ptfrac{\pibolth(a)}{\piref(a)} + \pibolth(a) \tfrac{1}{\pibolth(a)}}\\
\belowcomment{moving gradient inside and using identity $\myint{a} \gradth \pibolth(a)$=0}
&= \myint{a} \pibolth(a) \gradth \log\pibolth(a)\log \ptfrac{\pibolth(a)}{\piref(a)} \\
&= \Eb{a \sim \pibolth}{\gradth \log \pibolth(a) \log \ptfrac{\pibolth(a)}{\piref(a)}} \label{klgrad}
\end{align}
Continuing from \Cref{banditpart1end} but setting $\pi=\pibolth$,
\begin{align}
\gradth L_{\pi}(\qth) \evalat{\pi=\pibolth}
&= \Eb{a \sim \pibolth,r}{\tau \gradth \log\pibolth(a) \lrparen*{\vth - r} + \tau^2 \gradth \kl{\pibolth}{\piref}} \nonumber\\
&\qquad\quad\qquad+ \gradth \Eb{a \sim \pibolth, r}{\vth \lrparen*{\vth + \tau \kl{\pibolth}{\piref}-r}} \label{blk1}\\
&\hspace{-2em}= - \tau  \underbrace{\gradth\Eb{a \sim \pibolth, r}{r - \tau \kl{\pibolth}{\piref}}}_{\text{policy gradient}} + \underbrace{\gradth \Eb{a \sim \pi, r}{\half(\vth - ( r - \tau \kl{\pi}{\piref}))^2}}_{\text{value error gradient}}\evalat{\pi=\pibolth} \label{blk2}
\end{align}
Hence, the gradient of the squared error for our action-value function can be broken into two parts: the first part is the policy gradient of the Boltzmann policy corresponding to $\qth$, the second part arises from a squared error objective, where we are fitting $\vth$ to the entropy-augmented expected reward $\barr(a) - \temp \kl{\pibolth}{\piref}$.

Soon we will derive an equivalent interpretation of $Q$-function regression in the MDP setting, where we are approximating the state-value function $\Qpigam$.
However, we first need to introduce an entropy-regularized version of the reinforcement learning problem.

\section{Entropy-Regularized Reinforcement Learning} \label{secerrl}

We shall consider an entropy-regularized version of the reinforcement learning problem, following various prior work (\cite{ziebart2010modeling,fox2015taming,haarnoja2017reinforcement,nachum2017bridging}).
Specifically, let us define the \textit{entropy-augmented return} to be $\sum_{t=0}^{\infty} \gamma^t (r_t - \temp \K_t)$ where $r_t$ is the reward, $\gamma \in [0,1]$ is the discount factor, $\temp$ is a scalar temperature coefficient, and $\K_t$ is the Kullback-Leibler divergence between the current policy $\pi$ and a reference policy $\piref$ at timestep $t$: $\K_t = \kl{\pi(\cdot \given s_t)}{\piref(\cdot \given s_t)}$.
We will sometimes use the notation $\K(s) = \kl{\pi}{\piref}(s) = \kl{\pi(\cdot \given s)}{\piref(\cdot \given s)}$.
To emulate the effect of a standard entropy bonus (up to a constant), one can define $\piref$ to be the uniform distribution.
The subsequent sections will generalize some of the concepts from reinforcement learning to the setting where we are maximizing the entropy-augmented discounted return.

\subsection{Value Functions}
We are obliged to alter our definitions of value functions to include the new KL penalty terms.
We shall define the state-value function as the expected return:
\begin{align}
\Vpi(s) = \Ea{\sum_{t=0}^{\infty} \gamma^t (r_t - \temp \K_t) \givenb s_0=s} \label{vpidef}
\end{align}
and we shall define the $Q$-function as
\begin{align}
\Qpi(s, a) = \Ea{r_0 + \sum_{t=1}^{\infty} \gamma^t (r_t - \temp \K_t) \givenb s_0=s, a_0=a} \label{qpidef}
\end{align}
Note that this $Q$-function does not include the first KL penalty term, which does not depend on the action $a_0$.
This definition makes some later expressions simpler, and it leads to the following relationship between $\Qpi$ and $\Vpi$:
\begin{align}
\Vpi(s)
&= \Eb{a \sim \pi}{\Qpi(s,a)} - \temp \K(s),
\end{align}
which follows from matching terms in the sums in \Cref{vpidef,qpidef}.
\subsection{Boltzmann Policy}
In standard reinforcement learning, the ``greedy policy'' for $Q$ is defined as $[\G Q](s)=\argmax_a Q(s,a)$.
With entropy regularization, we need to alter our notion of a greedy policy, as the optimal policy is stochastic.
Since $\Qpi$ omits the first entropy term, it is natural to define the following stochastic policy, which is called the Boltzmann policy, and is analogous to the greedy policy:
\begin{align}
\pibolq(\cdot \given s)
&= \argmax_{\pi} \lrbrace*{ \Eb{a \sim \pi}{Q(s,a)} - \tau \kl{\pi}{\piref}(s)}\\
&= \piref(a \given s) \exp(Q(s,a)/\tau) / \underbrace{\Eb{a' \sim \piref}{\exp(Q(s,a')/\tau)}}_{\text{normalizing constant}}.
\end{align}
where the second equation is analogous to \Cref{pistardef} from the bandit setting.

Also analogously to the bandit setting, it is natural to define $V_Q$ (a function of $Q$) as
\begin{align}
V_Q(s) = \tau \log \Eb{a' \sim \piref}{\exp(Q(s,a')/\tau)}
\end{align}
so that
\begin{align}
\pibolq(a \given s) = \piref(a \given s) \exp((Q(s,a) - V_Q(s))/\tau) \label{boltz-def}
\end{align}
Under this definition, it also holds that
\begin{align}
V_Q(s) = \Eb{a \sim \pibolq(s)}{Q(s,a)} - \tau \kl{\pibolq}{\piref}(s) \label{mdpvid}
\end{align}
in analogy with \Cref{banditvthid}.
Hence, $V_Q(s)$ can be interpreted as an estimate of the expected entropy-augmented return, under the Boltzmann policy $\pibolq$.

Another way to interpret the Boltzmann policy is as the exponentiated advantage function.
Defining the advantage function as $A_Q(s,a) = Q(s,a) - V_Q(s)$, \Cref{boltz-def} implies that $\frac{\pibolq(a \given s)}{\piref(a \given s)} = \exp(A_Q(s,a)/\tau)$.

\subsection{Fixed-Policy Backup Operators}

The $\Tpi$ operators (for $Q$ and $V$) in standard reinforcement learning correspond to computing the expected return with a one-step lookahead: they take the expectation over one step of dynamics, and then fall back on the value function at the next timestep.
We can easily generalize these operators to the entropy-regularized setting.
We define
\begin{align}
[\Tpi V](s) &= \Eb{a\sim \pi,(r, s') \sim P(r, s' \given s, a)}{r - \tau \K(s) + \gamma V(s')} \label{vtpi}\\
[\Tpi Q](s, a) &= \Eb{(r, s') \sim P(r, s' \given s, a)}{r + \gamma (\Eb{a' \sim \pi}{Q(s',a')} - \tau \K(s'))}. \label{qtpi}
\end{align}

Repeatedly applying the $\Tpi$ operator $(\Tpi^n V = \underbrace{\Tpi(\Tpi(\dots \Tpi}_{\text{$n$ times}}(V))))$ corresponds to computing the expected return with a multi-step lookahead. That is, repeatedly expanding the definition of $\Tpi$, we obtain
\begin{align}
&[\Tpi^n V](s)=\Ea{\sum_{t=0}^{n-1}\gamma^t (r_t - \tau \K_t)  + \gamma^n V(s_n) \givenb s_0=s}\\
&[\Tpi^n Q](s,a) - \tau \K(s) = \Ea{\sum_{t=0}^{n-1} \gamma^t (r_t - \tau \K_t)  + \gamma^n (Q(s_n, a_n) - \tau \K_n) \givenb s_0=s, a_0=a}. \label{qtpin}
\end{align}
As a sanity check, note that in both equations, the left-hand side and right-hand side correspond to estimates of the total discounted return $\sum_{t=0}^{\infty} \gamma^t (r_t - \tau \K_t)$.

The right-hand side of these backup formulas can be rewritten using ``Bellman error'' terms $\delta_t$.
To rewrite the state-value ($V$) backup, define
\begin{align}
\delta_t = (r_t - \tau\K_t) + \gamma V(s_{t+1}) - V(s_t) \label{deltadef}
\end{align}
Then we have
\begin{align}
[\Tpi^n V](s)=\Ea{\sum_{t=0}^{n-1}\gamma^t \delta_t  + \gamma^n V(s_n) \givenb s_0=s}.
\end{align}

\subsection{Boltzmann Backups}
We can define another set of backup operators corresponding to the Boltzmann policy, $\pi(a \given s) \propto \piref(a \given s) \exp(Q(s,a)/\tau)$.
We define the following Boltzmann backup operator:
\begin{align}
[\T Q](s,a) &= \Eb{(r,s') \sim P(r,s' \given s,a)}{r + \gamma \underbrace{\Eb{a' \sim \G Q}{Q(s,a)} - \tau \kl{\G Q}{\piref}(s')}_{(\ast)}}\\
&=\Eb{(r,s') \sim P(r,s' \given s,a)}{r + \gamma \underbrace{\tau \log \Eb{a'\sim\piref}{\exp(Q(s',a')/\tau)}}_{(\ast\ast)}}
\end{align}
where the simplification from $(\ast)$ to $(\ast\ast)$ follows from the same calculation that we performed in the bandit setting (\Cref{banditvdef,banditvthid}).

The $n$-step operator $\Tpi^n$ for $Q$-functions also simplifies in the case that we are executing the Boltzmann policy.
Starting with the equation for $\Tpi^n Q$ (\Cref{qtpin}) and setting $\pi=\pibolq$, and then using \Cref{mdpvid} to rewrite the expected $Q$-function terms in terms of $V_Q$, we obtain
\begin{align}
[(\T_{\pibolq})^n Q](s,a) - \tau \K(s)
&= \Ea{\sum_{t=0}^{n-1} \gamma^t (r_t - \tau \K_t)  + \gamma^n (Q(s_n, a_n) - \tau \K_n) \givenb s_0=s, a_0=a}\\
&= \Ea{\sum_{t=0}^{n-1} \gamma^t (r_t - \tau \K_t)  + \gamma^n V_Q(s_n) \givenb s_0=s, a_0=a}.
\end{align}
From now on, let's denote this $n$-step backup operator by $\T_{\pibol, n}$. (Note $T_{\pibolq,n} \neq \T^n Q$, even though $\T_{\pibolq,1}Q=\T Q$, because $\T_{\pibolq}$ depends on $Q$.)

One can similarly define the \tdlam{} version of this backup operator
\newcommand{\Tpibol}{\T_{\pibolq}}
\newcommand{\Tpibollam}{\T_{\pibolq, \lambda}}
\newcommand{\Tpiboln}{\T_{\pibolq, n}}
\begin{align}
[\Tpibollam Q] = (1-\lambda)(1 + \lambda \Tpibol + (\lambda \Tpibol)^2 + \dots) \Tpibol Q. \label{tbollam}
\end{align}
One can straightforwardly verify by comparing terms that it satisfies
\begin{align}
&[\Tpibollam Q](s,a) = Q(s,a) + \Ea{\sum_{t=0}^{\infty} (\gamma \lambda)^t \delta_t \givenb s_0=s, a_0=a},\nonumber\\
&\where \delta_t = (r_t - \temp \K_t) + \gamma V_Q(s_{t+1}) - V_Q(s_t).
\label{eq:qlambackup}
\end{align}

\subsection{Soft \titleQ-Learning} \label{softq}
The Boltzmann backup operators defined in the preceding section can be used to define practical variants of $Q$-learning that can be used with nonlinear function approximation.
These methods, which optimize the entropy-augmented return, will be called soft $Q$-learning.
Following \cite{mnih2015human}, modern implementations of $Q$-learning, and $n$-step $Q$-learning (see \cite{mnih2016asynchronous}) update the $Q$-function incrementally to compute the backup against a fixed target $Q$-function, which we'll call $\Qtarg$.
In the interval between each target network update, the algorithm is approximately performing the backup operation $Q \leftarrow \T \Qtarg$ ($1$-step) or $Q \leftarrow \T_{\pibol_{\Qtarg},n} \Qtarg$ ($n$-step). To perform this approximate minimization, the algorithms minimize the least squares loss
\begin{alignat}{2}
L(Q) &= \Eb{t, s_t,a_t}{\half (Q(s_t, a_t) - y_t)^2}, \where \\
y_t &= r_t + \gamma V_{\Qtarg}(s_{t+1}) \qquad &&\text{1-step $Q$-learning} \label{1step}\\
y_t &= \tau \K_t + \sum_{d=0}^{n-1} \gamma^d (r_{t+d} - \tau \K_{t+d}) + \gamma^{n} V_{\Qtarg}(s_{t+n}) \qquad &&\text{$n$-step $Q$-learning}    \label{nstep} \\
&= \tau \K_t + V_{\Qtarg}(s_t) + \sum_{d=0}^{n-1} \gamma^d \delta_{t+d} \nonumber \\
&\where \delta_t = (r_t - \tau \K_t) + \gamma V_{\Qtarg}(s_{t+1}) - V_{\Qtarg}(s_t) \label{nstepq-target}
\end{alignat}
In one-step $Q$-learning (\Cref{1step}), $y_t$ is an unbiased estimator of $[\T Q](s_t, a_t)$, regardless of what behavior policy was used to collect the data.
In $n$-step $Q$-learning (\Cref{nstep}), for $n>1$, $y_t$ is only an unbiased estimator of $[\T_{\pibol_{\Qtarg},n} \Qtarg](s_t, a_t)$ if actions $a_t, a_{t+1}, \dots, a_{t+d-1}$ are sampled using $\pibol_{\Qtarg}$.
\subsection{Policy Gradients} \label{entpg}

Entropy regularization is often used in policy gradient algorithms, with gradient estimators of the form
\begin{align}
\Eb{t, s_t, a_t}{\gradth \log \pith(a_t \given s_t) \sum_{t' \ge t} r_{t'} - \temp \gradth \kl{\pith}{\piref}(s_t)} \label{naivepg}
\end{align}
(\cite{williams1992simple,mnih2016asynchronous}).

However, these are not proper estimators of the entropy-augmented return $\sum_{t} (r_t - \temp \K_t)$, since they don't account for how actions affect entropy at future timesteps.
Intuitively, one can think of the KL terms as a cost for ``mental effort''. \Cref{naivepg} only accounts for the instantaneous effect of actions on mental effort, not delayed effects.

To compute proper gradient estimators, we need to include the entropy terms in the return.
We will define the discounted policy gradient in the following two equivalent ways---first, in terms of the empirical return; second, in terms of the value functions $\Vpi$ and $\Qpi$:
\begin{align}
g_{\gamma}(\pith)
&= \Ea{\sum_{t=0}^{\infty} \gradth \log \pith(a_t \given s_t) \lrparen*{r_0 + \sum_{d=1}^{\infty} \gamma^{d} (r_{t+d}-\temp\K_{t+d}) - \temp \gradth \kl{\pith}{\piref}(s_t)}} \\
&= \Ea{\sum_{t=0}^{\infty} \gradth \log \pith(a_t \given s_t) (\Qpi(s_t, a_t)- \Vpi(s_t) ) - \temp \gradth \kl{\pith}{\piref}(s_t)}
\end{align}
\begin{minipage}{.5\textwidth}
In the special case of a finite-horizon problem---i.e., $r_t=\K_t=0$ for all $t \ge T$---the undiscounted ($\gamma=1$) return is finite, and it is meaningful to compute its gradient.
In this case, $g_1(\pith)$ equals the undiscounted policy gradient:
\begin{align}
g_{1}(\pi) = \gradth \Ea{\sum_{t=0}^{T-1}(r_t - \temp \K_t)}
\end{align}
This result is obtained directly by considering the stochastic computation graph for the loss (\cite{schulman2015gradient}), shown in the figure on the right. The edges from $\theta$ to the KL loss terms lead to the $\gradth \kl{\pith}{\piref}(s_t)$ terms in the gradient; the edges to the stochastic actions $a_t$ lead to the $\gradth \log \pith(a_t \given s_t) \sum_{t=d}^{T-1} (r_{t+d} - \tau \K_{t+d})$ terms in the gradient.
\end{minipage}
\begin{minipage}{.5\textwidth}
\begin{center}
\includegraphics[height=.8\textwidth]{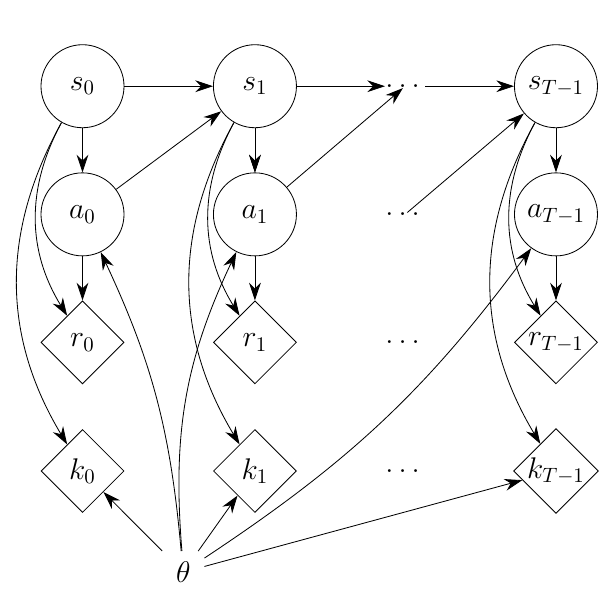}
\end{center}
\end{minipage}

Since $g_1(\pith)$ computes the gradient of the entropy-regularized return, one interpretation of $\ggam(\pith)$ is that it is an approximation of the undiscounted policy gradient $g_1(\pith)$, but that it allows for lower-variance gradient estimators by ignoring some long-term dependencies.
A different interpretation of $\ggam(\pi)$ is that it gives a gradient flow such that $\pistar=\pibol_{\Qstar}$ is the (possibly unique) fixed point.

As in the standard MDP setting, one can define approximations to $\ggam$ that use a value function to truncate the returns for variance reduction.
These approximations can take the form of $n$-step methods (\cite{mnih2016asynchronous}) or \tdlam{}-like methods (\cite{schulman2015high}), though we will focus on $n$-step returns here.
Based on the definition of $\ggam$ above, the natural choice of variance-reduced estimator is
\begin{align}
\Eb{t, s_t, a_t}{\gradth \log \pith(a_t \given s_t) \sum_{d=0}^{n-1} \gamma^d \delta_{t+d} } \label{nsteppg}
\end{align}
where $\delta_t$ was defined in \Cref{deltadef}.

The state-value function $V$ we use in the above formulas should approximate the entropy augmented return $\sum_{t=0}^{\infty}\gamma^t (r_t - \tau \K_t)$.
We can fit $V$ iteratively by approximating the $n$-step backup $V \leftarrow \Tpi^n V$, by minimizing a squared-error loss
\begin{align}
&L(V) = \Eb{t, s_t}{\half (V(s_t) - y_t)^2},\\
&\where y_t = \sum_{d=0}^{n-1} \gamma^d r_{t+d} + \gamma^d V(s_{t+d}) = V(s_t) + \sum_{d=0}^{n-1} \gamma^d \delta_{t+d}. \label{vtarg}
\end{align}

\section{Soft \titleQ-learning Gradient Equals Policy Gradient} \label{segpg}

This section shows that the gradient of the squared-error loss from soft $Q$-learning (\Cref{softq}) equals the policy gradient (in the family of policy gradients described in \Cref{entpg}) plus the gradient of a squared-error term for fitting the value function.
We will not make any assumption about the parameterization of the $Q$-function, but we define $\Vth$ and $\pith$ as the following functions of the parameterized $Q$-function $\Qth$:
\begin{align}
\Vth(s) &\defeq \tau \log \Eb{a}{\exp(\Qth(s,a)/\temp)}\\
\pith(a \given s) &\defeq \piref(a \given s) \exp((\Qth(s,a)-\Vth(s))/\temp)
\end{align}
Here, $\pith$ is the Boltzmann policy for $\Qth$, and $\Vth$ is the normalizing factor we described above.
From these definitions, it follows that the $Q$-function can be written as
\begin{align}
\Qth(s,a) = \Vth(s) + \temp \log \tfrac{\pith(a \given s)}{\piref(a \given s)} \label{qthident}
\end{align}
We will substitute this expression into the squared-error loss function.
First, for convenience, let us define $\Dlam_t = \sum_{d=0}^{n-1} \gamma^d \delta_{t+d}$.

Now, let's consider the gradient of the $n$-step soft $Q$-learning objective:
\begin{align}
&\gradth \Eb{t, s_t, a_t \sim \pi}{\half \norm*{\Qth(s_t,a_t) - y_t}^2}\evalat{\pi=\pith}\\
\belowcomment{swap gradient and expectation, treating state-action distribution as fixed:}
&= \Eb{t, s_t, a_t \sim \pi}{\gradth \Qth(s_t,a_t) (\Qth(s_t,a_t) - y_t)}\evalat{\pi=\pith}\\
\belowcomment{replace $\Qth$ using \Cref{qthident}, and replace $Q$-value backup $y_t$ by \Cref{nstep}:}
&= \E_{t, s_t, a_t \sim \pi} \Big[ \gradth \Qth(s_t,a_t) (\temp \log \pifrac + \Vth(s_t) - (\Vth(s_t) + \temp \relent(s_t) + \Dlam_t)) \Big]\evalat{\pi=\pith}\\
\belowcomment{cancel out $\Vth(s_t)$:}
&= \E_{t, s_t, a_t \sim \pi} \Big[ \gradth \Qth(s_t,a_t) (\temp \log \pifrac  - \temp \relent(s_t) - \Dlam_t)\Big]\evalat{\pi=\pith} \\
\belowcomment{replace the other $\Qth$ by \Cref{qthident}:}
&= E_{t, s_t, a_t \sim \pi}\Big[(\temp \gradth \log \pith(a_t \given s_t) + \gradth \Vth(s_t))
\cdot (\temp \log \pifrac - \temp \relent(s_t)-\Dlam_t)\Big]\evalat{\pi=\pith}\\
\belowcomment{expand out terms:}
&= E_{t, s_t, a_t \sim \pi}\Big[
\temp^2\gradth \log \pith(a_t \given s_t) \log \tfrac{\pith(a_t \given s_t)}{\piref(a_t \given s_t)}
- \temp^2 \underbrace{\gradth \log \pith(a_t \given s_t) \relent(s_t)}_{(\ast)} \nonumber \\
& - \temp \gradth \log \pith(a_t \given s_t) \Dlam_t
+\underbrace{\temp\gradth \Vth(s_t) \log \tfrac{\pith(a_t \given s_t)}{\piref(a_t \given s_t)}- \temp \gradth \Vth(s_t) \relent(s_t)}_{(\ast\ast)}
-  \gradth \Vth(s_t) \Dlam_t
\Big]\evalat{\pi=\pith}\\
\belowcomment{($\ast$) vanishes because $\Eb{a \sim \pith(\cdot \given s_t)}{\gradth \log \pith(a_t \given s_t)\cdot const}=0$}
\belowcomment{($\ast\ast$) vanishes because $\Eb{a \sim \pith(\cdot \given s_t)}{ \tfrac{\pith(a_t \given s_t)}{\piref(a_t \given s_t)}  }=\relent(s_t)$}
&= E_{t, s_t, a_t \sim \pi}\Big[
\temp^2 \gradth \relent(s_t)
+ 0 - \temp \gradth \log \pith(a_t \given s_t) \Dlam_t + 0
-  \gradth \Vth(s_t) \Dlam_t
\Big]\evalat{\pi=\pith}\\
\belowcomment{rearrange terms:}
&= \E_{t, s_t, a_t \sim \pi} \Big[\underbrace{-\temp \gradth \log \pith(a_t \given s_t) \Dlam_t   + \temp^2 \gradth \relent](s_t)}_{\text{policy grad}}
+ \underbrace{\gradth \half\norm*{\Vth(s_t) - \hatv_t}^2}_{\text{value function grad}}\Big]\evalat{\pi=\pith} \label{equiv-eqn}
\end{align}
Note that the equivalent policy gradient method multiplies the policy gradient by a factor of $\tau$, relative to the value function error.
Effectively, the value function error has a coefficient of $\tau^{-1}$, which is larger than what is typically used in practice (\cite{mnih2016asynchronous}). We will analyze this choice of coefficient in the experiments.

\section{Soft \titleQ-learning and Natural Policy Gradients} \label{secnpg}

The previous section gave a first-order view on the equivalence between policy gradients and soft $Q$-learning; this section gives a second-order, coordinate-free view.
As previous work has pointed out, the natural gradient is the solution to a regression problem; here we will explore the relation between that problem and the nonlinear regression in soft $Q$-learning.

The natural gradient is defined as $F^{-1} g$, where $F$ is the average Fisher information matrix, $F=\Eb{s,a \sim \pi}{(\gradth \log \pith(a \given s))^T (\gradth \log \pith(a \given s)) }$, and $g$ is the policy gradient estimate $g \propto \Ea{\gradth \log \pith(a \given s) \Dlam}$, where $\Dlam$ is an estimate of the advantage function.
As pointed out by \cite{kakade2002natural}, the natural gradient step can be computed as the solution to a least squares problem.
Given timesteps $t=1,2,\dots,T$, define $\bpsi_t = \gradth \log\pith(a_t \given s_t)$.
Define $\bPsi$ as the matrix whose $t^{\text{th}}$ row is $\bpsi_t$, let $\bm{\Delta}$ denote the vector whose $t^{\text{th}}$ element is the advantage estimate $\Delta_t$, and let $\epsilon$ denote a scalar stepsize parameter.
Consider the least squares problem
\begin{align}
\min_{\bw} \half \norm*{ \bPsi \bw - \epsilon \bm{\Delta} }^2 \label{ls}
\end{align}
The least-squares solution is $\bw = \epsilon (\bPsi^T \bPsi)^{-1} \bPsi^T \bm{\Delta}$.
Note that $\Ea{\bPsi^T \bPsi}$ is the Fisher information matrix $F$, and $\Ea{\bPsi^T \bm{\Delta}}$ is the policy gradient $g$, so $\bw$ is the estimated natural gradient.

Now let us interpret the least-squares problem in \Cref{ls}. $\bPsi \bw$ is the vector whose $t^{\text{th}}$ row is $\gradth \log \pith(a \given s) \cdot \bw$.
According to the definition of the gradient,  if we perform a parameter update with $\theta - \thold = \epsilon\bw$, the change in $\log \pith(a \given s)$ is as follows, to first order in $\epsilon$:
\begin{align}
\log \pith(a \given s) - \log \pithold(a \given s) \approx \gradth \log \pith(a \given s) \cdot \epsilon \bw = \epsilon \bpsi \cdot \bw
\end{align}
Thus, we can interpret the least squares problem (\Cref{ls}) as solving
\begin{align}
\min_{\theta} \sum_{t=1}^T \half ( \log \pith(a_t \given s_t) - \log \pithold(a_t \given s_t) - \epsilon \Delta_t )^2
\label{natgrad-ls}
\end{align}
That is, we are adjusting each log-probility $\log \pithold(a_t \given s_t)$ by the advantage function $\Delta_t$, scaled by $\epsilon$.

In entropy-regularized reinforcement learning, we have an additional term for the gradient of the KL-divergence:
\begin{align}
g
&\propto \Ea{\gradth \log \pith(a_t \given s_t) \Delta_t  - \tau\gradth \K[\pith,\piref](s_t)} \\
&= \Ea{\gradth \log \pith(a_t \given s_t) \lrparen*{\Delta_t  - \tau \lrbrack*{\log \ptfrac{\pith(a_t \given s_t)}{\piref(a_t \given s_t)} - \K[\pith, \piref](s_t)} }}
\end{align}
where the second line used the formula for the KL-divergence  (\Cref{klgrad}) and the identity that \\ $\Eb{a_t\sim\pith}{\gradth \log \pith(a_t \given s_t) \cdot \mathrm{const}}=0$ (where the KL term is the constant.)
In this case, the corresponding least squares problem (to compute $F^{-1}g$) is
\begin{align}
\min_{\theta} \sum_{t=1}^T \half \lrparen*{ \log \pith(a_t \given s_t) - \log \pithold(a_t \given s_t) - \epsilon\lrparen*{ \Delta_t - \tau \lrbrack*{\log \ptfrac{\pith(a_t \given s_t)}{\piref(a_t \given s_t)} - \K[\pithold, \piref](s_t)} }}^2.
\label{natgrad-ls2}
\end{align}

Now let's consider $Q$-learning. Let's assume that the value function is unchanged by optimization, so $\Vth = \Vthold$. (Otherwise, the equivalence will not hold, since the value function will try to explain the measured advantage $\Dlam$, shrinking the advantage update.)
\begin{align}
\half \lrparen*{\Qth(s_t, a_t) - y_t}^2
&= \half \lrparen*{ \lrparen*{\Vth(s_t, a_t) + \temp \log \ptfrac{\pith(a_t \given s_t)}{\piref(a_t \given s_t)}} - \lrparen*{\Vthold(s_t) + \temp \K[\pithold,\piref](s_t) + \Dlam_t}}^2 \\
&= \half \lrparen*{\temp \log \ptfrac{\pith(a_t \given s_t)}{\piref(a_t \given s_t)} - (\Dlam_t + \temp \K[\pithold,\piref](s_t))}^2
\end{align}
Evidently, we are regressing $\log \pith(a_t \given s_t)$ towards $\log\pithold(a_t \given s_t) + \Dlam_t/\temp + \K[\pithold, \piref](s_t)$.
This loss is \textit{not} equivalent to the natural policy gradient loss that we obtained above.

We can recover the natural policy gradient by instead solving a \textit{damped} version of the $Q$-function regression problem. Define $\hatqeps_t = (1-\epsilon) \Qthold(s_t, a_t) + \epsilon \hatq_t$, i.e., we are interpolating between the old value and the backed-up value.
\begin{align}
&\hatqeps_t = (1-\epsilon) \Qthold(s_t, a_t) + \epsilon \hatq_t = \Qthold(s_t, a_t) + \epsilon (\hatq_t - \Qthold(s_t, a_t)) \\
&\hatq_t - \Qthold(s_t, a_t) = \lrparen*{\Vth(s_t) + \temp \K[\pithold, \piref](s_t) + \Dlam_t} - \lrparen*{\Vthold(s_t) + \temp \log \ptfrac{\pithold(a_t \given s_t)}{\piref(a_t \given s_t)}} \\
&= \Dlam_t + \temp \lrbrack*{\K[\pithold, \piref](s_t) - \log \ptfrac{\pithold(a_t \given s_t)}{\piref(a_t \given s_t)}}\\
&\Qth(s_t, a_t) - \hatqeps_t
= \Qth(s_t, a_t) - \lrparen*{\Qthold(s_t, a_t) + \epsilon \lrparen*{\hatq_t - \Qthold(s_t, a_t) }}\\
&=   \Vth(s_t) + \log \ptfrac{\pith(a_t \given s_t)}{\piref(a_t \given s_t)}  - \lrbrace*{\Vthold(s_t) + \log \ptfrac{\pithold(a_t \given s_t)}{\piref(a_t \given s_t)}  + \epsilon\lrparen*{\Dlam + \temp\lrbrack*{\K[\pithold, \piref](s_t) - \log \ptfrac{\pithold(a_t \given s_t)}{\piref(a_t \given s_t)}}}}  \nonumber \\
& =  \log \pith(a_t \given s_t) - \log \pithold(a_t \given s_t) - \epsilon \lrparen*{\Dlam_t - \temp\lrbrack*{\log \ptfrac{\pithold(a_t \given s_t)}{\piref(a_t \given s_t)} - \K[\pithold, \piref](s_t)}}
\end{align}
which exactly matches the expression in the least squares problem in \Cref{natgrad-ls2}, corresponding to entropy-regularized natural policy gradient.
Hence, the ``damped'' $Q$-learning update corresponds to a natural gradient step.

\section{Experiments}\label{secexp}
To complement our theoretical analyses, we designed experiments to study the following questions:
\begin{enumerate}
\item Though one-step entropy bonuses are used in PG methods for neural network policies  (\cite{williams1992simple,mnih2016asynchronous}), how do the entropy-regularized RL versions of policy gradients and $Q$-learning described in \Cref{secerrl} perform on challenging RL benchmark problems? How does the ``proper'' entropy-regularized policy gradient method (with entropy in the returns) compare to the naive one (with one-step entropy bonus)? (\Cref{a2catari})
\item How do the entropy-regularized versions of $Q$-learning (with logsumexp) compare to the standard DQN of \cite{mnih2015human}?  (\Cref{dqnatari})
\item The equivalence between PG and soft $Q$-learning is established \textit{in expectation}, however, the actual gradient estimators are slightly different due to sampling.
Furthermore, soft $Q$-learning is equivalent to PG with a particular penalty coefficient on the value function error. Does the equivalence hold under practical conditions? (\Cref{entretpg})
\end{enumerate}

\subsection{A2C on Atari: Naive vs Proper Entropy Bonuses}\label{a2catari}

Here we investigated whether there is an empirical effect of including entropy terms when computing returns, as described in \Cref{secerrl}.
In this section, we compare the naive and proper policy gradient estimators:
\begin{align}
& \text{naive / 1-step:} \qquad \grad \log \pith(a_t \given s_t) \lrparen*{\sum_{d=0}^{n-1} \gamma^d r_{t+d} -V(s_t)} - \tau \gradth \relent(s_t) \\
& \text{proper:} \qquad \grad \log \pith(a_t \given s_t) \lrparen*{\sum_{d=0}^{n-1} \gamma^d (r_{t+d} - \tau \relent(s_{t+d})) - V(s_t)} - \tau \gradth \relent(s_t)
\end{align}
In the experiments on Atari, we take $\piref$ to be the uniform distribution, which gives a standard entropy bonus up to a constant.

We start with a well-tuned (synchronous, deterministic) version of A3C (\cite{mnih2016asynchronous}), henceforth called A2C (advantage actor critic), to optimize the entropy-regularized return.
We use the parameter $\tau = 0.01$ and train for $320$ million frames.
We did not tune any hyperparameters for the ``proper'' algorithm---we used the same hyperparameters that had been tuned for the ``naive'' algorithm.

\begin{figure}[!h]
\centering
\includegraphics[width=.3\linewidth]{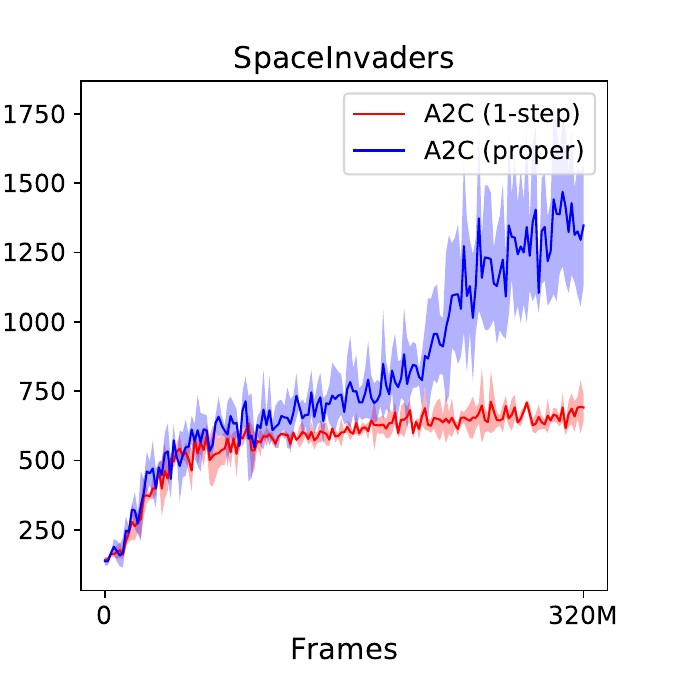}
\includegraphics[width=.3\linewidth]{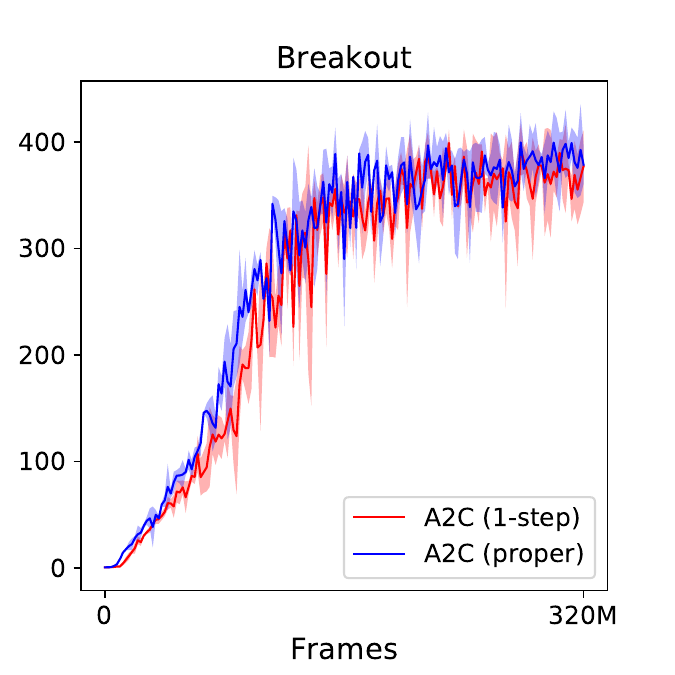}
\includegraphics[width=.3\linewidth]{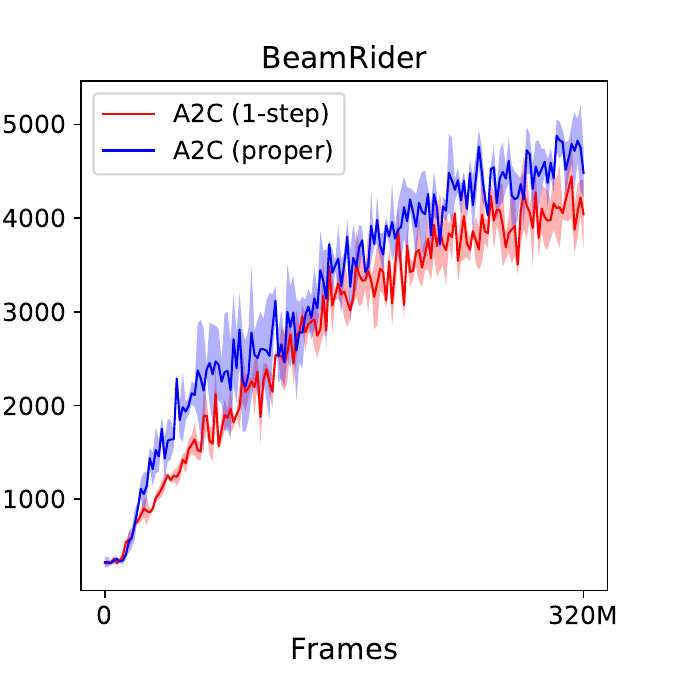}
\includegraphics[width=.3\linewidth]{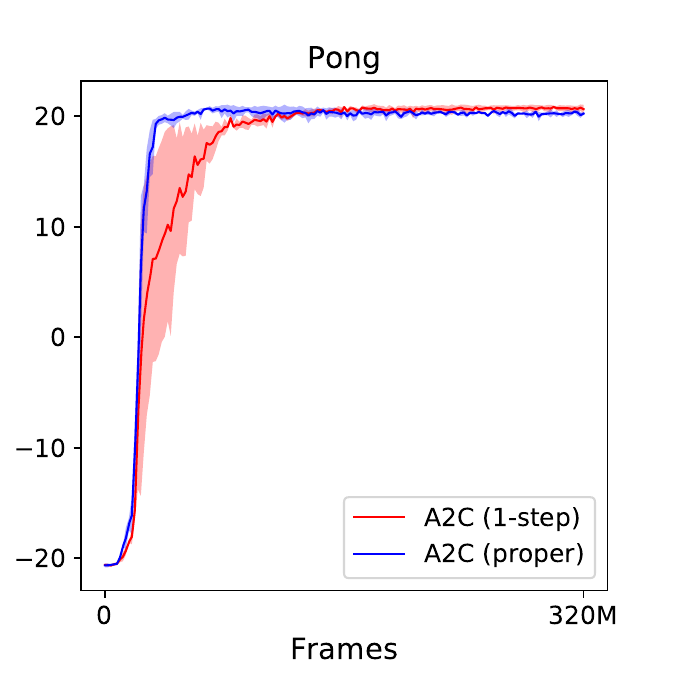}
\includegraphics[width=.3\linewidth]{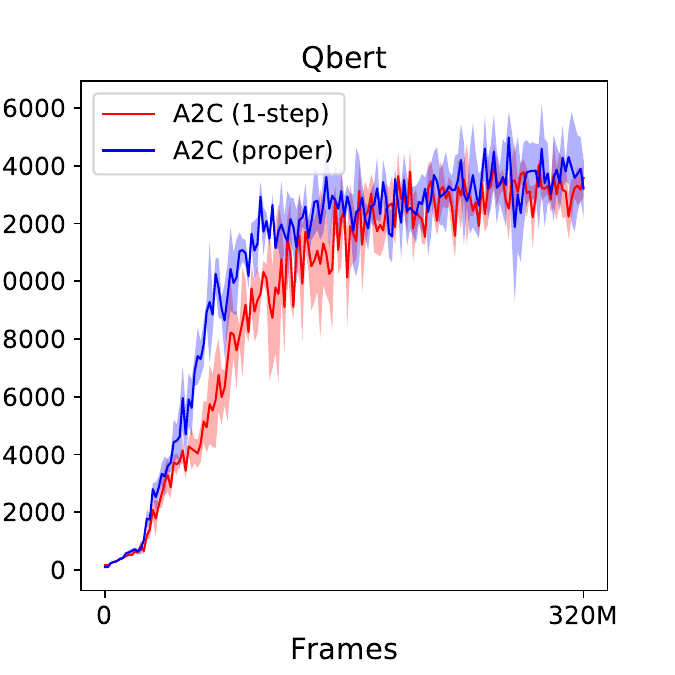}
\includegraphics[width=.3\linewidth]{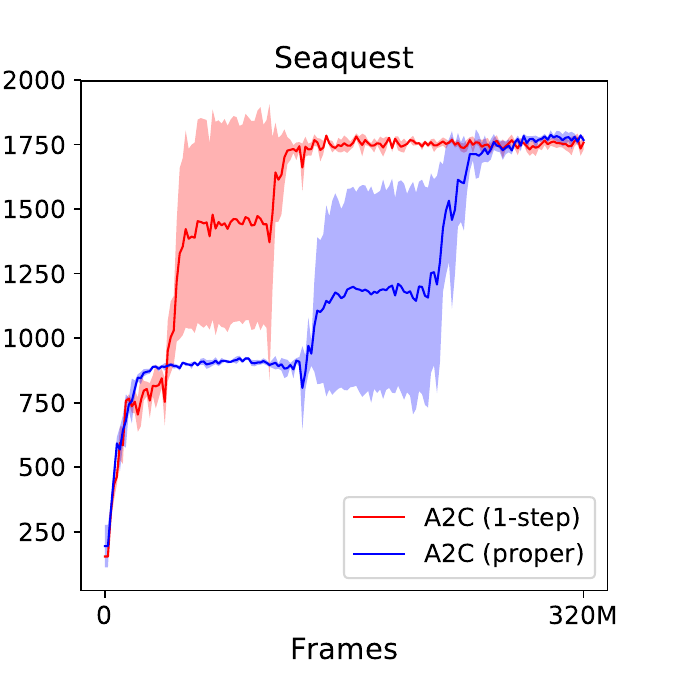}
\caption{Atari performance with different RL objectives. EntRL is A2C modified to optimize for return augmented with entropy (instead of KL penalty). Solid lines are average evaluation return over $3$ random seeds and shaded area is one standard deviation.}\label{fig:atari}
\end{figure}

As shown in \Cref{fig:atari}, the ``proper'' version yields performance that is the same or possibly greater than the ``naive'' version.
Hence, besides being attractive theoretically, the entropy-regularized formulation could lead to practical performance gains.

\subsection{DQN on Atari: Standard vs Soft}\label{dqnatari}

Here we investigated whether soft $Q$-learning (which optimizes the entropy-augmented return) performs differently from standard ``hard'' $Q$-learning on Atari.
We made a one-line change to a DQN implementation:
\begin{alignat}{2}
&y_t = r_t + \gamma \max_{a'} Q(s_{t+1}, a')  &&\qquad\text{Standard}\\
&y_t = r_t + \gamma \log \sum_{a'} \exp(Q(s_{t+1}, a')/\tau)-\log\abs{\mathcal{A}}  &&\qquad\text{``Soft'': KL penalty}\\
&y_t =r_t + \gamma \log \sum_{a'} \exp(Q(s_{t+1}, a')/\tau)  &&\qquad\text{``Soft'': Entropy bonus}
\end{alignat}
The difference between the entropy bonus and KL penalty (against uniform) is simply a constant, however, this constant made a big difference in the experiments, since a positive constant added to the reward encourages longer episodes. Note that we use the same epsilon-greedy exploration in all conditions; the only difference is the backup equation used for computing $y_t$ and defining the loss function.

\begin{figure}
\centering
\includegraphics[width=.3\linewidth]{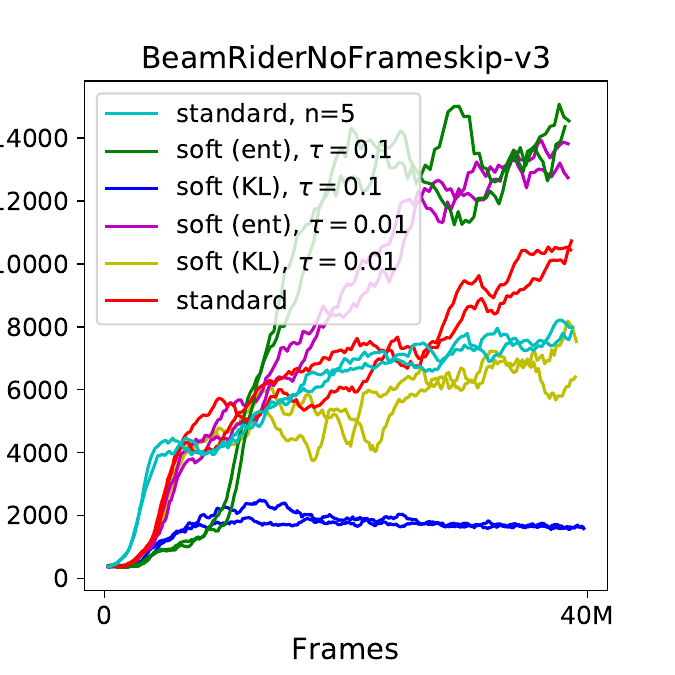}\hspace{2em}
\includegraphics[width=.3\linewidth]{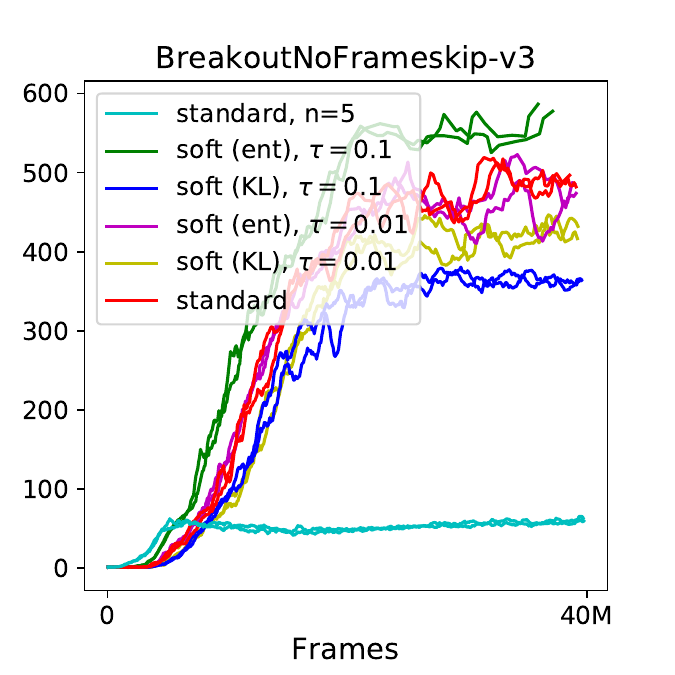}\hspace{2em}
\includegraphics[width=.3\linewidth]{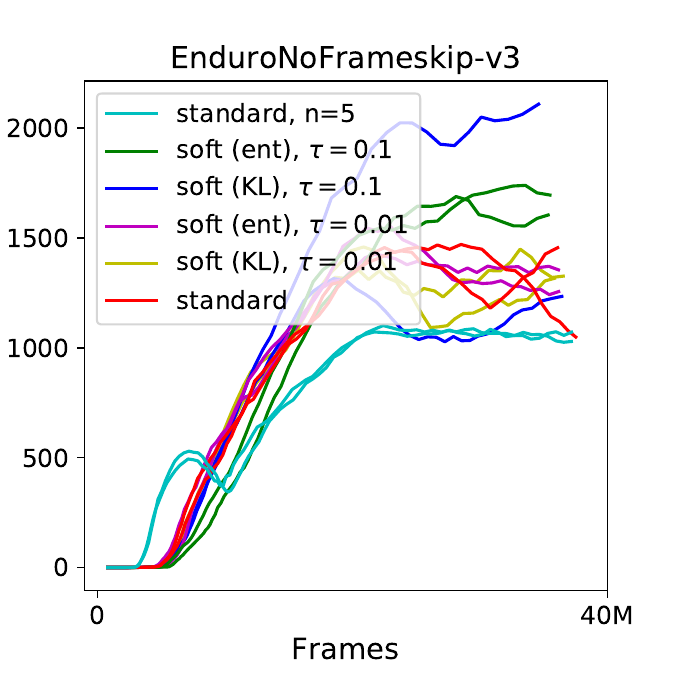}\\
\includegraphics[width=.3\linewidth]{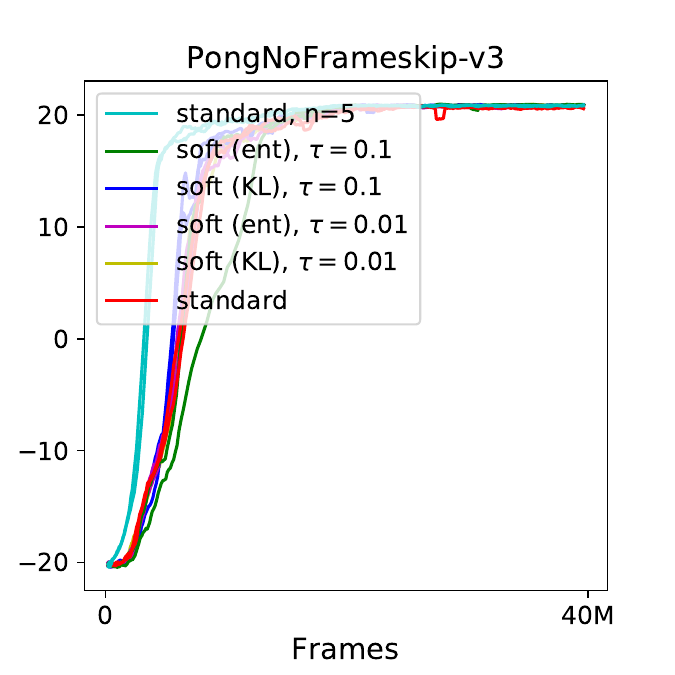}\hspace{2em}
\includegraphics[width=.3\linewidth]{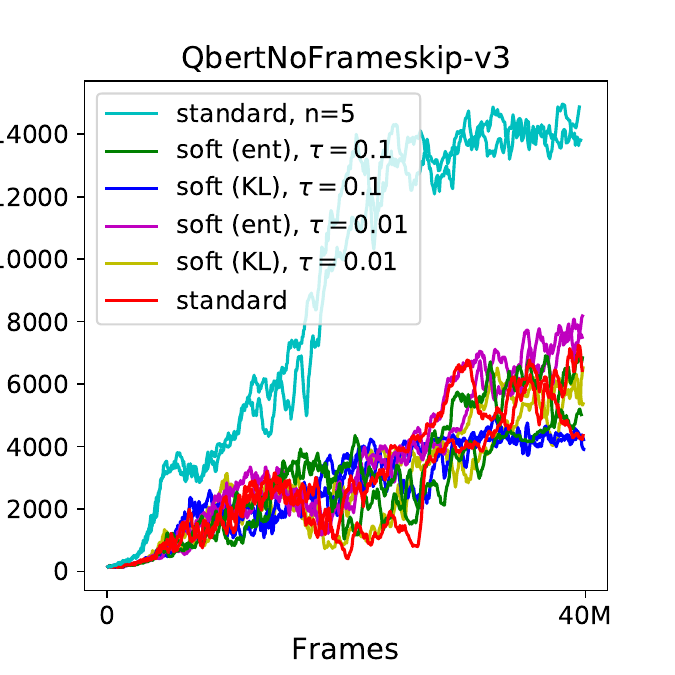}\hspace{2em}
\includegraphics[width=.3\linewidth]{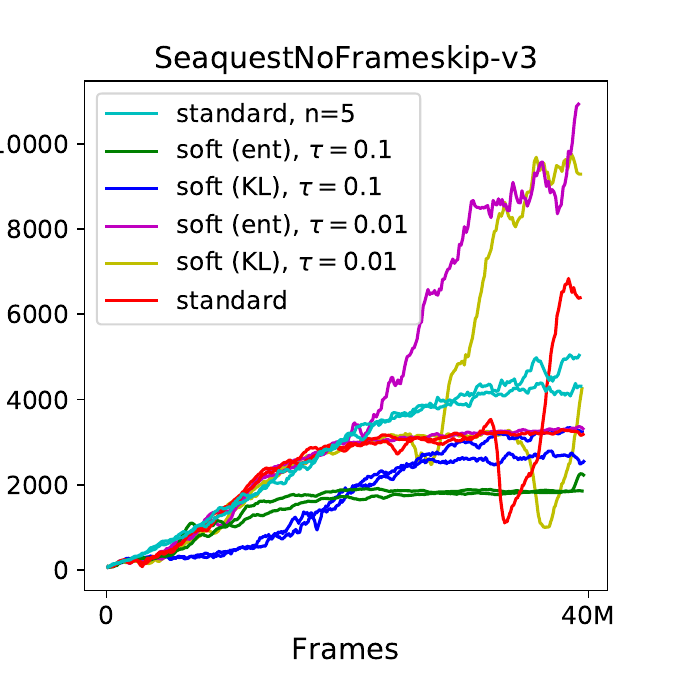}
\caption{Different variants of soft $Q$-learning and standard $Q$-learning, applied to Atari games. Note that 4 frames = 1 timestep. \label{softdqn}}
\end{figure}

The results of two runs on each game are shown in \Cref{softdqn}. The entropy-bonus version with $\tau=0.1$ seems to perform a bit better than standard DQN, however, the KL-bonus version performs worse, so the benefit may be due to the effect of adding a small constant to the reward. We have also shown the results for $5$-step $Q$-learning, where the algorithm is otherwise the same. The performance is better on Pong and $Q$-bert but worse on other games---this is the same pattern of performance found with $n$-step policy gradients. (E.g., see the A2C results in the preceding section.)

\subsection{Entropy Regularized PG vs Online \titleQ-Learning on Atari} \label{entretpg}

Next we investigate if the equivalence between soft $Q$-learning and PG is relevant in practice---we showed above that the gradients are the same in expectation, but their variance might be different, causing different learning dynamics. For these experiments,  we modified the gradient update rule used in A2C while making no changes to any algorithmic component, i.e. parallel rollouts, updating parameters every $5$ steps, etc.
The $Q$-function was represented as: $\Qth(s,a) = \Vth(s) + \temp \log \pith(a \given s)$, which can be seen as a form of dueling architecture with $\temp \log \pith(a \given s)$ being the ``advantage stream'' (\cite{wang2015dueling}).
$\Vth, \pith$ are parametrized as the same neural network as A2C, where convolutional layers and the first fully connected layer are shared.
$\pith(a \given s)$ is used as behavior policy.

A2C can be seen as optimizing a combination of a policy surrogate loss and a value function loss, weighted by hyperparameter $c$:
\begin{align}
& L_\mathrm{policy} = - \log \pith(a_t \given s_t) \Dlam_t   + \temp \relent](s_t) \\
& L_\mathrm{value} = \half\norm*{\Vth(s_t) - \hatv_t}^2 \\
& L_\mathrm{a2c} = L_\mathrm{policy} + c L_\mathrm{value}
\end{align}
In normal A2C, we have found $c=0.5$ to be a robust setting that works across multiple environments.
On the other hand, our theory suggests that if we use this $Q$-function parametrization, soft $Q$-learning has the same expected gradient as entropy-regularized A2C with a specific weighting $c = \frac{1}{\temp}$.
Hence, for the usual entropy bonus coefficient setting $\temp = 0.01$, soft $Q$-learning is implicitly weighting value function loss a lot more than usual A2C setup ($c = 100$ versus $c = 0.5$).
We have found that such emphasis on value function ($c = 100$) results in unstable learning for both soft $Q$-learning and entropy-regularized A2C.
Therefore, to make $Q$-learning exactly match known good hyperparameters used in A2C, we scale gradients that go into advantage stream by $\frac{1}{\gamma}$ and scale gradients that go into value function stream by $c = 0.5$.

\begin{figure}[!h]
\centering
\includegraphics[width=.3\linewidth]{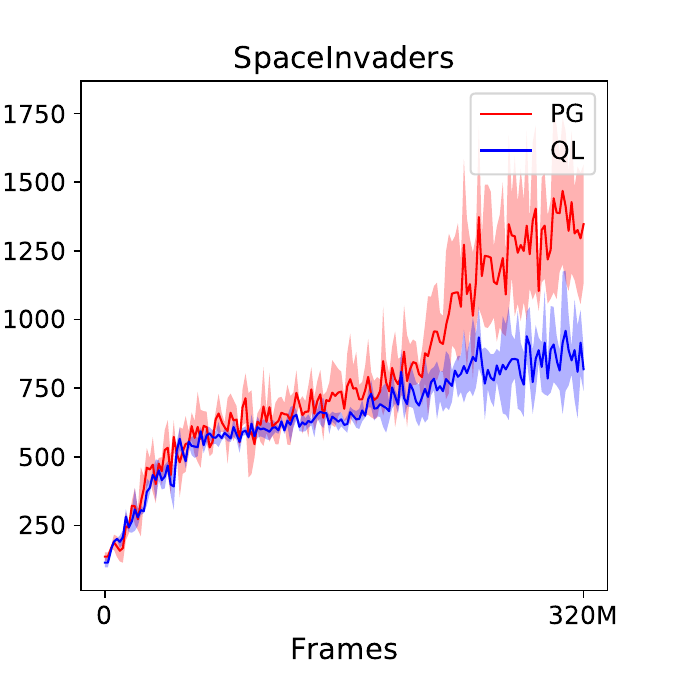}
\includegraphics[width=.3\linewidth]{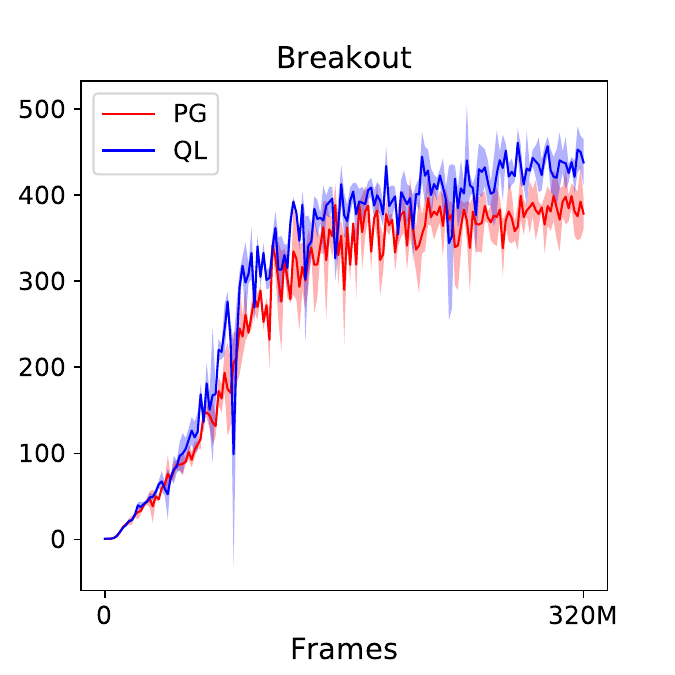}
\includegraphics[width=.3\linewidth]{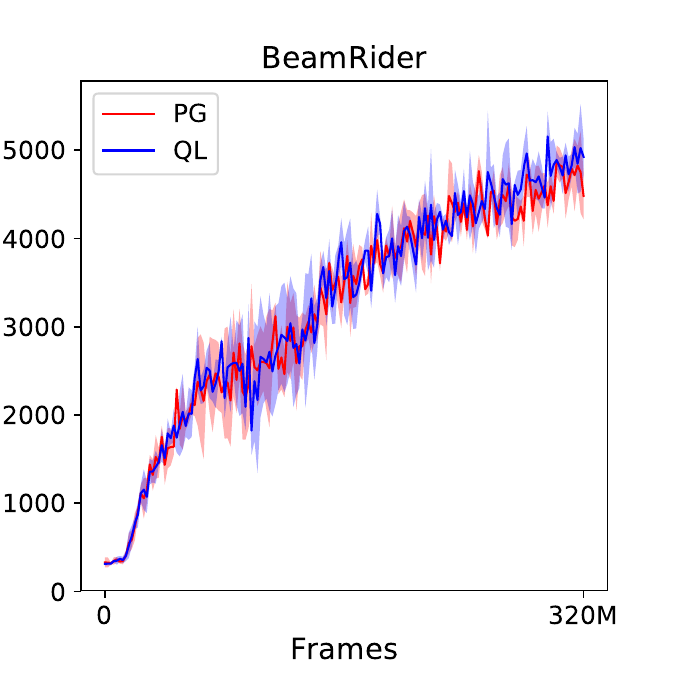}
\includegraphics[width=.3\linewidth]{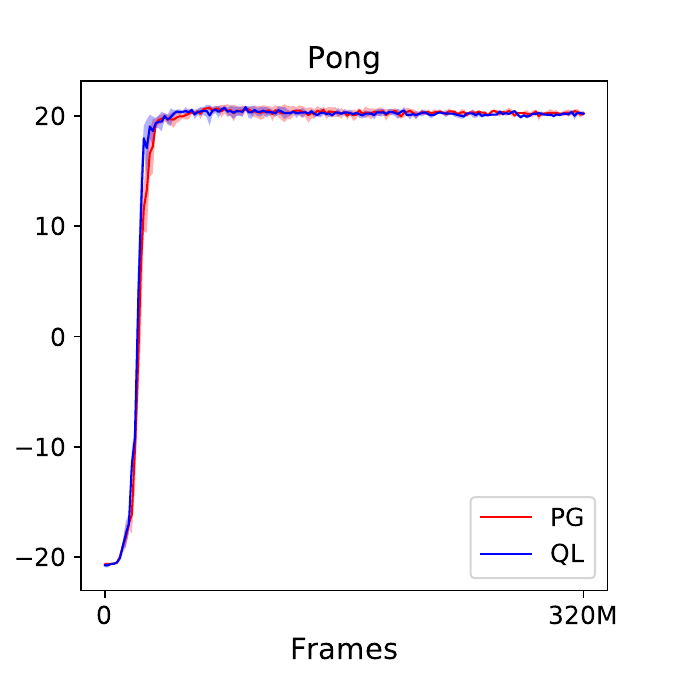}
\includegraphics[width=.3\linewidth]{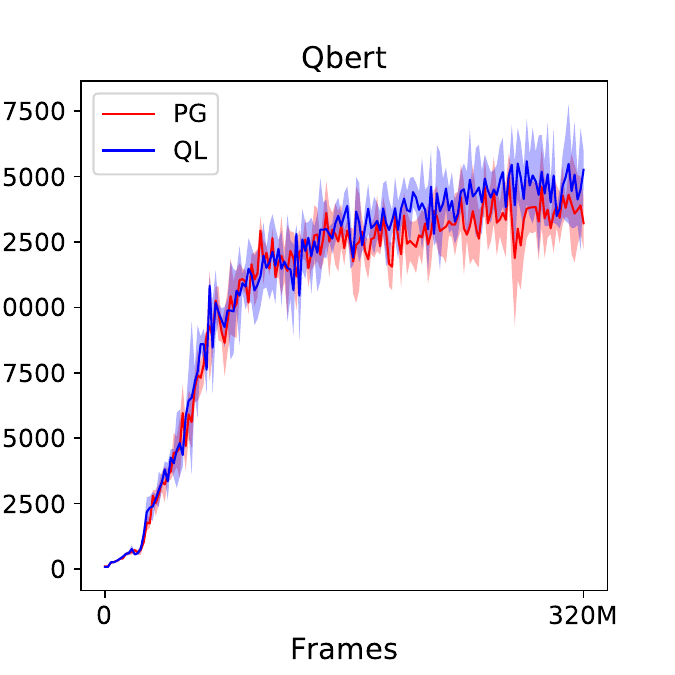}
\includegraphics[width=.3\linewidth]{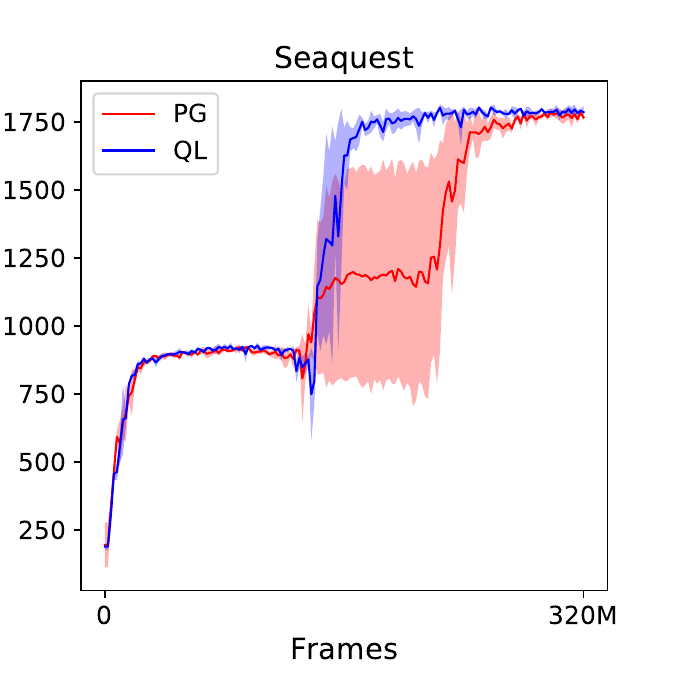}
\caption{Atari performance with policy gradient vs $Q$-learning update rules. Solid lines are average evaluation return over $3$ random seeds and shaded area is one standard deviation.}\label{fig:atari2}
\end{figure}

With the same default A2C hyperparameters, learning curves of PG and QL are almost identical in most games (\Cref{fig:atari2}), which indicates that the learning dynamics of both update rules are essentially the same even when the gradients are approximated with a small number of samples.
Notably, the $Q$-learning method here demonstrates stable learning without the use of target network or $\epsilon$ schedule.

\section{Related Work}
Three recent papers have drawn the connection between policy-based methods and value-based methods, which becomes close with entropy regularization.
\begin{itemize}
\item \cite{o2016pgq} begin with a similar motivation as the current paper: that a possible explanation for  $Q$-learning and SARSA is that their updates are similar to policy gradient updates.
They decompose the $Q$-function into a policy part and a value part, inspired by dueling $Q$-networks (\cite{wang2015dueling}):
\begin{align}
Q(s,a) = V(s) + \tau(\log \pi(a \given s) + \tau S[\pi(\cdot \given s)])
\end{align}
This form is chosen so that the term multiplying $\tau$ has expectation zero under $\pi$, which is a property that the true advantage function satisfies: $\Eb{\pi}{\Api}=0$.
Note that our work omits that $S$ term, because it is most natural to define the $Q$-function to not include the first entropy term.
The authors show that taking the gradient of the Bellman error of the above $Q$-function leads to a result similar to the policy gradient.
They then propose an algorithm called PGQ that mixes together the updates from different prior algorithms.
\item \cite{nachum2017bridging} also discuss the entropy-regularized reinforcement learning setting, and develop an off-policy method that applies in this setting.
Their argument (modified to use our notation and KL penalty instead of entropy bonus) is as follows.
The advantage function $\Api(s,a) = \Qpi(s,a) - \Vpi(s)$ lets us define a multi-step consistency equation, which holds even if the actions were sampled from a different (suboptimal) policy.
In the setting of deterministic dynamics, $\Qpi(s_t,a_t) = r_t + \gamma\Vpi(s_{t+1})$,
hence
\begin{align}
\sum_{t=0}^{n-1} \gamma^t \Api(s_t, a_t) = \sum_{t=0}^{n-1} \gamma^t (r_t + \gamma \Vpi(s_{t+1}) - \Vpi(s_t)) = \sum_{t=0}^{n-1} \gamma^t r_t + \gamma^n \Vpi(s_n) - \Vpi(s_0) \label{multistepadvit}
\end{align}
If $\pi$ is the optimal policy (for the discounted, entropy-augmented return), then it is the Boltzmann policy for $\Qpi$, thus
\begin{align}
\tau (\log\pi(a \given s) - \log\piref(a \given s)) = A_{\Qpi}(s,a)
\end{align}
This expression for the advantage can be substituted into \Cref{multistepadvit}, giving the consistency equation
\begin{align}
\sum_{t=0}^{n-1} \gamma^t \tau (\log \pi(s_t, a_t) - \log\piref(s_t, a_t)) = \sum_{t=0}^{n-1} \gamma^t r_t + \gamma^n \Vpi(s_n) - \Vpi(s_0), \label{multistepthing2}
\end{align}
which holds when $\pi$ is optimal. The authors define a squared error objective formed from by taking LHS - RHS in \Cref{multistepthing2}, and jointly minimize it with respect to the parameters of $\pi$ and $V$.
The resulting algorithm is a kind of Bellman residual minimization---it optimizes with respect to the future target values, rather than treating them as fixed \cite{scherrer2010should}.
\item \cite{haarnoja2017reinforcement} work in the same setting of soft $Q$-learning as the current paper, and they are concerned with tasks with high-dimensional action spaces, where we would like to learn stochastic policies that are multi-modal, and we would like to use $Q$-functions for which there is no closed-form way of sampling from the Boltzmann distribution $\pi(a \given s) \propto \piref(a \given s) \exp(Q(s,a)/\tau)$. Hence, they  use a method called Stein Variational Gradient Descent to derive a procedure that jointly updates the $Q$-function and a policy $\pi$, which approximately samples from the Boltzmann distribution---this resembles variational inference, where one makes use of an approximate posterior distribution.
\end{itemize}

\section{Conclusion}
We study the connection between two of the leading families of RL algorithms used with deep neural networks. In a framework of entropy-regularized RL we show that soft $Q$-learning is equivalent to a policy gradient method (with value function fitting) in terms of expected gradients (first-order view). In addition, we also analyze how a damped $Q$-learning method can be interpreted as implementing natural policy gradient (second-order view).
Empirically, we show that the entropy regularized formulation considered in our theoretical analysis works in practice on the Atari RL benchmark, and that the equivalence holds in a practically relevant regime.

\section{Acknowledgements}
We would like to thank Matthieu Geist for pointing out an error in the first version of this manuscript, Chao Gao for pointing out several errors in the second version, and colleagues at OpenAI for insightful discussions.

\bibliographystyle{plainnat}
\bibliography{pg-qr.bib}

\end{document}

%% file: rl_math.tex
\DeclareMathOperator*{\argmax}{arg\,max}

\newcommand{\half}{\tfrac{1}{2}}
\newcommand{\defeq}{\vcentcolon=}
\newcommand{\given}{\:\vert\:}
\newcommand{\givenb}{\:\middle\vert\:}

\newcommand{\grad}{\nabla}
\newcommand{\evalat}[1]{\big\rvert_{#1}}

\newcommand{\E}{\mathbb{E}}

\newcommand{\Ea}[1]{\E\left[#1\right]}
\newcommand{\Eb}[2]{\E_{#1}\left[#2\right]}

\DeclarePairedDelimiter{\norm}{\|}{\|}
\DeclarePairedDelimiter{\abs}{|}{|}

\DeclarePairedDelimiter{\lrbrack}{[}{]}
\DeclarePairedDelimiter{\lrbrace}{ \{ }{ \} }
\DeclarePairedDelimiter{\lrparen}{(}{)}

\newcommand{\Qpi}{\ensuremath{Q_{\pi}}}
\newcommand{\Api}{\ensuremath{A_{\pi}}}
\newcommand{\Vpi}{\ensuremath{V_{\pi}}}
\newcommand{\Qstar}{\ensuremath{Q_{*}}}

\newcommand{\kl}[2]{D_{\mathrm{KL}}\left[#1 ~ \middle \| ~ #2\right]}

\newcommand{\tdlam}{TD($\lambda$)}
\newcommand{\tdzero}{TD($0$)}

\newcommand{\myint}[1]{\int \! \! \mathrm{d} #1 \ }